\documentclass{article}

\usepackage{spconf}

\usepackage{amsmath,amssymb}

\usepackage{graphicx}
\usepackage{subfig}      
\usepackage{caption}     
\usepackage{multirow}

\usepackage{cite}
\usepackage[hyphens]{url}
\usepackage[hidelinks]{hyperref}

\usepackage{listings}
\usepackage{verbatim}

\title{Scaling Audio-Visual Quality Assessment Dataset via Crowdsourcing}
\name{Renyu Yang$^{1}$, Jian Jin$^{1}$, Lili Meng$^{2}$, Meiqin Liu$^{3}$, Yilin Wang$^{4}$, Balu Adsumilli$^{4}$, Weisi Lin$^{1}$}
\address{$^1$College of Computing and Data Science, Nanyang Technological University, 639798, Singapore.\\$^2$School of Information Science and Engineering, Shandong Normal University, 250014, China\\$^3$Institute of Information Science, Beijing Jiao Tong University, 100044, China.\\$^4$YouTube Media Algorithms team, Google Inc., Mountain View, CA, 94043, USA}
\begin{document}
%
\maketitle
\begin{abstract}
Audio-visual quality assessment (AVQA) research has been stalled by limitations of existing datasets: they are typically small in scale, with insufficient diversity in content and quality, and annotated only with overall scores. These shortcomings provide limited support for model development and multimodal perception research. We propose a practical approach for AVQA dataset construction. First, we design a crowdsourced subjective experiment framework for AVQA, breaks the constraints of in-lab settings and achieves reliable annotation across varied environments. Second, a systematic data preparation strategy is further employed to ensure broad coverage of both quality levels and semantic scenarios. Third, we extend the dataset with additional annotations, enabling research on multimodal perception mechanisms and their relation to content. Finally, we validate this approach through YT-NTU-AVQ, the largest and most diverse AVQA dataset to date, consisting of 1,620 user-generated audio and video (A/V) sequences. The dataset and platform code are available at \href{https://github.com/renyu12/YT-NTU-AVQ}{https://github.com/renyu12/YT-NTU-AVQ}.
\end{abstract}

\section{Introduction}

Audio-Visual Quality Assessment (AVQA) provides essential guidance for optimizing multimedia quality by modeling the joint perceptual impact of auditory and visual signals \cite{Akhtar2017avqasurvey}. Unlike Video Quality Assessment (VQA) or Audio Quality Assessment (AQA), AVQA better reflects the comprehensive Quality of Experience (QoE) of subjects when they watch and listen to the A/V sequences.

As is known, data-driven paradigms have already shown remarkable success in signal quality assessment tasks \cite{zhang2025largemodelqa, wen2025cpllmcontextpixelaware}. However, AVQA research progress remains limited due to the lack of datasets. Existing AVQA datasets (representative comparisons in Table~\ref{tab:avqa_datasets}) remain scarce and generally small-scale~\cite{demirbilek2016inrs,martinez2014unbavq,min2020livesjtu,cao2023sjtuuav}. They are insufficient to support the training, fine-tuning, and evaluation of advanced data-driven models, let alone systematic studies of human multimodal perception.

The primary challenge lies in scaling AVQA subjective experiments, which typically require controlled in-lab environments and subjects with good auditory discrimination ability, severely limiting dataset size. In contrast, crowdsourcing provides a natural way to scale, but its less controlled environments are particularly problematic for AVQA, where both audio and video must be judged jointly~\cite{itu_p910_2023}. Nevertheless, prior subjective studies \cite{hosu2017konstanz,sinno2018livevqc,wang2019youtubeugc,ying2021lsvq,lin2022konjnd, wang2024youtube,chen2025mugsqa} have shown that appropriate crowdsourcing experimental design can improve annotation quality, raising the key question: \textit{Can AVQA subjective experiments be effectively scaled through crowdsourcing while maintaining annotation reliability in non-laboratory settings?}

In addition, existing AVQA datasets also suffer from a lack of diversity in terms of audio-visual content, distortion types, and quality distributions. Thus, they fail to capture the complexity of real-world user-generated content, which restricts their practical value and limits the generalizability of trained models. Furthermore, their annotations are typically limited to overall quality scores. It constrains the exploration of how different modalities and semantic factors shape human multimodal quality perception.

To address these 3 limitations, We propose a scalable approach for AVQA dataset building as follows:
\begin{itemize}
    \item We design the first crowdsourced subjective experiment framework for AVQA, enabling large-scale data annotation with reliable quality in non-laboratory environments through checks on experiment procedure, rating results and subjects. 
    \item We propose a stratified sampling and manual selection method to widely select distorted A/V sequences from open-access resources, enabling coverage of diverse distortion types and semantic scenarios. This results in YT-NTU-AVQ, the largest AVQA dataset to date of 1,620 user-generated A/V sequences.
    \item We extend the dataset with rich annotations, including modality-specific quality, attention weights, audio and video categories, and video summaries. This provides resources to study multimodal perception mechanisms and their dependence on semantic scenarios.
\end{itemize}

\begin{table*}[t]
\centering
\scriptsize
\begin{tabular}{|c|c|c|c|c|c|c|c|c|c|c|c|}
\hline
\textbf{Dataset} & \textbf{Source} & \textbf{Collection} & \textbf{Unique AV} & \textbf{Total AV} & \textbf{Distortion} & \textbf{Resolution} & \textbf{Length} & \textbf{Label} & \textbf{Rating Method} & \textbf{Year} \\
\hline
UnB-AVQ & CDVL & Manual & 6 & 72 & Synthetic & 720p & 8s & Multi-MOS & In-Lab & 2013 \\
LIVE-SJTU & CDVL & Manual & 14 & 336 & Synthetic & 1080p & 8s & AVQA MOS & In-Lab & 2019 \\
SJTU-UAV & YFCC100M & Manual & 520 & 520 & In-the-wild & 720p–1080p & 8s & AVQA MOS & In-Lab & 2022 \\
Ours & YouTube CC & Sampling + Manual & 1620 & 1620 & In-the-wild & $\leq$1080p & 10s$^{\dagger}$ & Multi-label & Crowdsourcing & 2025 \\
\hline
\end{tabular}
\caption{Comparison of representative AVQA datasets. ($^{\dagger}$A small fraction of YT-NTU-AVQ sequences are shorter than 10s)}
\label{tab:avqa_datasets}
\end{table*}

\section{Subjective Experiment Framework}

Crowdsourcing-based AVQA experiments provide an efficient alternative to traditional lab-based studies by removing physical constraints and enabling large-scale participation from diverse environments. However, AVQA-specific crowdsourcing remain largely unexplored due to unique challenges:

\begin{itemize}
    \item Varied experimental conditions: Subjects use various devices with different display and audio characteristics, which cannot be fully regulated.
    \item Inconsistent scoring standards: Subjects may interpret the rating task differently without intensive training.
    \item Unreliable submissions: A proportion of annotations may be careless, random, or even fraudulent.
\end{itemize}

Despite these issues, we argue that a carefully designed crowdsourcing experiment is both feasible and valuable for AVQA, because a large and diverse subject pool reduces random variability and improves the representativeness of collected opinions. Our goal is to capture general user perceptions under realistic conditions.

In addition, although these challenges cannot be fully resolved, they can be mitigated: basic environment control address varied experimental conditions, simple training reduces inconsistent interpretations of the task, appropriate filtering removes unreliable submissions, and subjects screening ensures the inclusion of reliable subjects.

Therefore, we develop a practical framework including a series of measures across experimental procedure, result filtering, and subject screening to improve the reliability and efficiency of subjective annotation. 

\subsection{Experiment Platform Design}
we developed a custom platform based on jsPsych~\cite{de2015jspsych}. It can support the complete experiment session: environment check, consent, instruction, training, and quality rating. Key features include: (1) enforcing environment requirements and playback restrictions, (2) logging user interactions to detect inattentive behavior, (3) providing brief instructions and five training videos, and (4) collecting multiple subjective annotations.

For environment control, we adopt methods such as soft confirmation (quiet space, proper screen, headphones), limited automatic checks (device, resolution, audio output, network), and enforced playback restrictions (full-screen mode, no timeline dragging, mandatory completion before submission).

To study multimodal perception, we collect multidimensional ratings rather than a single overall score. In a crowdsourcing setting, however, subjects’ attention is limited, so the number and complexity of questions must be carefully controlled. 4 questions in our experiment are as follows (single-stimulus with discretized slider ratings: Q1--Q3: 1.0--5.0, step 0.1; Q4: 0--100\%, step 1\%):
\begin{itemize}
    \item \textbf{Q1 (AVQA Score) :} \textit{Rate overall audio-visual quality.}
    \item \textbf{Q2 (AV\_VQA Score) :} \textit{Rate video quality only.}
    \item \textbf{Q3 (AV\_AQA Score) :} \textit{Rate audio quality only.}
    \item \textbf{Q4 (Weight of AV) :} \textit{Which Part do you pay more attention to when you evaluate the overall quality? (Give a 100\% split, e.g., Audio 50\% : 50\% Video).}
\end{itemize}

\begin{figure*}[t]
  \centering
  \includegraphics[width=0.32\textwidth, trim=0 15 0 20]{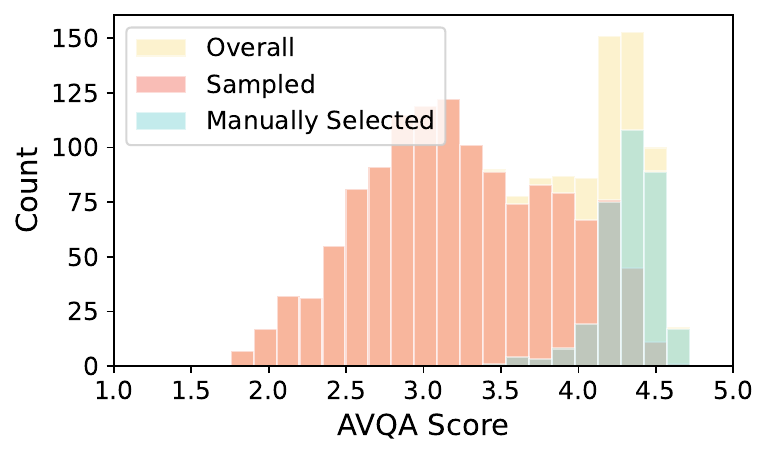}
  \includegraphics[width=0.32\textwidth, trim=0 15 0 20]{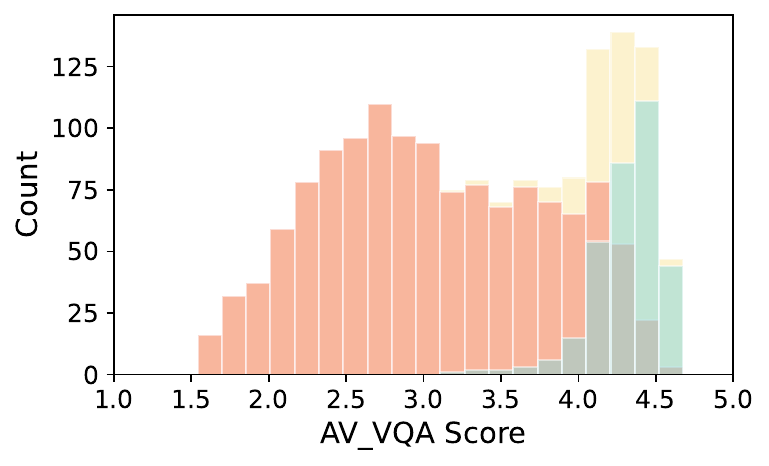}
  \includegraphics[width=0.32\textwidth, trim=0 15 0 20]{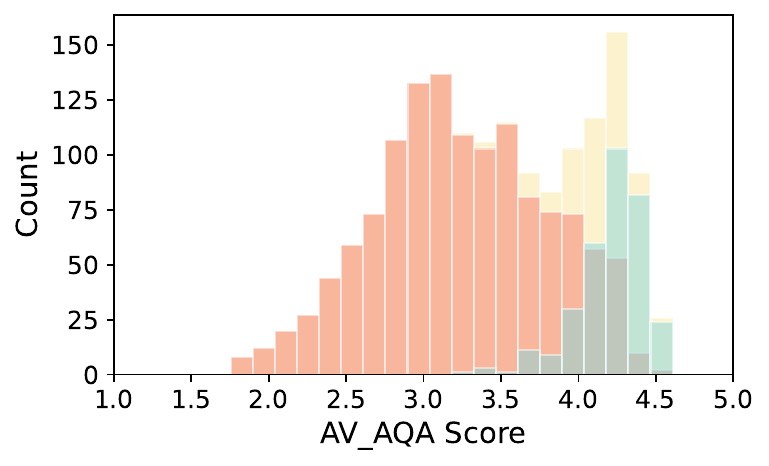}
  \caption{Distributions of MOS scores in our AVQA dataset. From left to right: overall audio-visual quality (AVQA), video-only quality (AV\_VQA), and audio-only quality (AV\_AQA). (Mean and standard deviation: AVQA: $\mu=3.47$, $\sigma=0.72$; AV\_VQA: $\mu=3.49$, $\sigma=0.77$; AV\_AQA: $\mu=3.44$, $\sigma=0.64$)}
  \label{fig:mos_distribution}
\end{figure*}

\subsection{Ranking-based Data Filtering}

A pilot study was conducted to validate the feasibility of our crowdsourcing design, but it also revealed a non-negligible amount of low-quality results. We tested different conventional filtering methods, but still struggled to identify a large portion of unreliable submissions, which often appeared as randomly distributed mid-range ratings.

Motivated by these failure cases, we propose a simple yet effective dynamic filtering approach combining \textbf{ranking consistency} (SROCC with group consensus) and \textbf{score dispersion} (standard deviation). Ranking consistency filters out randomized responses that break the overall ordering of videos, while score dispersion removes overly uniform or concentrated ratings. Only results exceeding both thresholds are considered reliable. This joint criterion reliably identifies uninformative or careless submissions, substantially improving rating quality.

\subsection{Multi-stage Experiment}
Relying solely on data filtering after the experiment is inefficient and costly, since many unreliable submissions may already be collected. We observed that submission quality strongly depends on subject reliability, motivating a three-stage framework to screen subjects.

In brief, we first build a small reference subset through the pretest, then dynamically screen new subjects in the qualification test, and finally let only qualified subjects rate the remaining set of videos in the formal test.  This design enables dynamic and scalable subject screening, ensuring that large-scale data collection remains both reliable and efficient.

The details in our experiment are as follows. \textbf{Pretest:} 120 videos are used to establish reference scores and identify reliable subjects via ranking-based filtering. \textbf{Qualification test:} new subjects rate 30 videos from the pretest subset and are qualified only if they meet the filtering criteria. \textbf{Formal test:} only qualified subjects proceed to rate the remaining 1,500 videos. Each group contains 30 videos to keep an acceptable session duration, and all results are ultimately rechecked with ranking-based filtering to discard low-quality data.

\section{Data Preparation}

Dataset diversity is essential for the generalization of AVQA models. Yet existing AVQA datasets rely on manual selection, which severely limits diversity and scalability. A better strategy is to first construct a large candidate pool and then apply stratified sampling, with limited manual selection used for supplementation. But a major challenge is \textbf{audio-relevant sequence selection}. Unlike VQA, where clips can be randomly cropped, AVQA requires segments with meaningful or salient audio, which are unevenly distributed and often sparse.

A feasible strategy is to combine audio event detection models with manual checking, while we adopt a simpler solution by leveraging the existing audio-visual understanding dataset VALOR~\cite{chen2023valor}, which already provides over 1M 10-second YouTube A/V sequences with rich audio-visual content and semantic labels.

From this pool, we applied a stratified sampling strategy to maintain coverage across multiple factors. Three attributes were prioritized: (1) \textbf{audio quality}, using pseudo-labels from AudioBox~\cite{tjandra2025audiobox}; (2) \textbf{video quality}, predicted by FasterVQA~\cite{wu2023fastervqa}; and (3) \textbf{audio semantics}, obtained by coarsely mapping AudioSet \cite{gemmeke2017audioset} labels into speech, music, and sound with seven combinations. We first sampled 10k candidates, then selected 1,296 while reweighting attribute distributions to broaden label coverage.

Another practical issue is that many sequences from existing sources\cite{cdvl2019,thomee2015yfcc100m} are outdated, with most videos published before 2015. To address this, we manually added 324 CC-licensed sequences from post-2017 YouTube videos, ensuring recency and topical diversity. Together, these steps yield a dataset of 1,620 clips with balanced coverage of semantic classes and quality ranges.

In addition, we complemented missing semantic annotations for all sequences with Gemini2.5 \cite{comanici2025gemini2.5} predictions and manual verification, covering audio and video categories, and video summaries.

\section{Results and Analysis}

\begin{figure}[t]
  \centering
  \includegraphics[width=\linewidth]{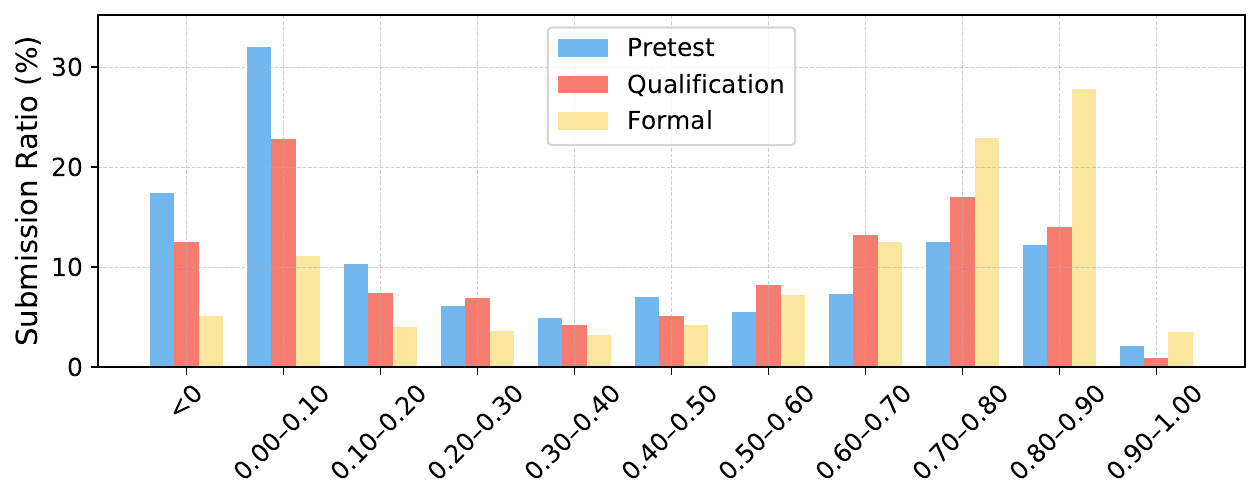}
  \caption{Distribution of average SROCC values across submissions in different crowdsourcing stages. Submission quality improves progressively from pretest to formal evaluation, validating the effectiveness of our multi-stage filtering pipeline.}
  \label{fig:srocc_distribution}
\end{figure}

\subsection{Subjective Experiment Results}

All experiments were conducted on Amazon Mechanical Turk (AMT) with subjects required to have more than 500 completed tasks and an approval rate above 97\%. After applying filtering criteria ($SROCC > 0.5$, $STD > 0.5$), the three stages yielded the following:

\textbf{Pretest.} 328 tasks for 120 videos produced 9,840 ratings from 286 subjects, of which 3,750 (31.3 per video, 38.1\% acceptance) were retained as reliable references. \textbf{Qualification.} 1,952 new participants rated one pretest group each, yielding 32,490 valid ratings (270.8 per video, 49.1\% acceptance) and qualifying 1,051 subjects. \textbf{Formal.} For the remaining 1,500 videos, 4,680 tasks produced 94,590 reliable ratings from 580 qualified subjects (63.1 per video, 67.4\% acceptance).

Figure \ref{fig:mos_distribution} presents the MOS distributions of the overall AVQA score, video-only quality score (AV\_VQA), and audio-only quality score (AV\_AQA). The sampled subset exhibits approximately Gaussian-shaped quality distributions, indicating the effectiveness of our stratified sampling. The manually selected subset further supplements the high-quality range, ensuring broader coverage of quality.

Figure~\ref{fig:srocc_distribution} shows clear improvements in submission reliability across stages, confirming the effectiveness of our experiment framework.

\begin{figure}[t]
  \centering
  \subfloat[]{\includegraphics[width=0.32\linewidth, trim=0 15 0 40]{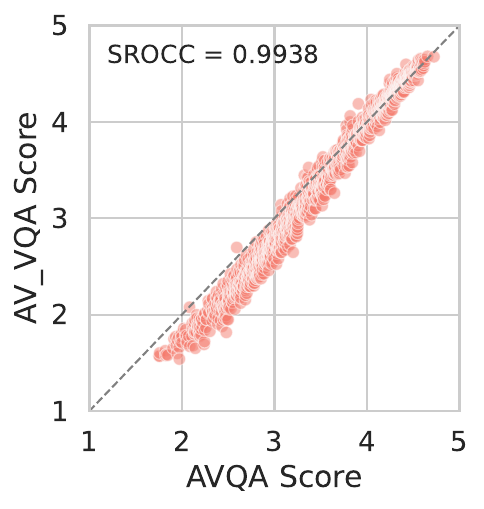}}
  \hfill
  \subfloat[]{\includegraphics[width=0.32\linewidth, trim=0 15 0 40]{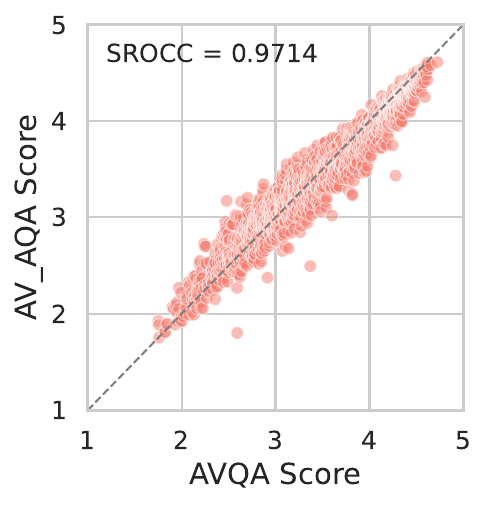}}
  \hfill
  \subfloat[]{\includegraphics[width=0.32\linewidth, trim=0 15 0 40]{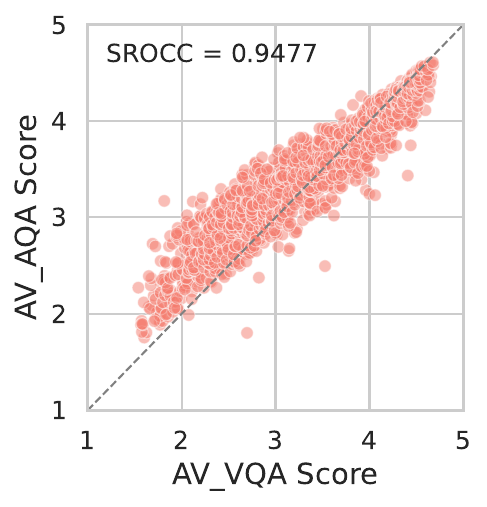}}
  \caption{Scatter plots of MOS comparisons: (a) AVQA vs AV\_VQA, (b) AVQA vs AV\_AQA, and (c) AV\_VQA vs AV\_AQA. All three pairs show high correlations, revealing strong monotonic relationships among the score types.}
  \label{fig:score_scatter_plots}
\end{figure}


\begin{table}[t]
\centering
\renewcommand{\arraystretch}{1.1}
\setlength{\tabcolsep}{5pt}
\begin{tabular}{|c|l|cc|}
\hline
\textbf{Type} & \textbf{Method} & \textbf{SROCC} & \textbf{PLCC} \\
\hline

\multirow{3}{*}{AQA}
 & PAM & 0.4217 & 0.4358 \\
 & AudioBox (CE) & 0.4328 & 0.4335 \\
 & AudioBox (PQ) & 0.5145 & 0.5335 \\
\hline

\multirow{3}{*}{VQA}
 & FasterVQA & 0.7868 & 0.7865 \\
 & FastVQA & 0.8503 & 0.8514 \\
 & Q-Align & 0.9342 & 0.9329 \\
\hline

\multirow{5}{*}{AVQA}
 & BRISQUE+MFCC & 0.4078 & 0.4233 \\
 & V-BLIINDS+eGeMAPS & 0.6525 & 0.6630 \\
 & VRN-50+ARN-50 & 0.8425 & 0.8412 \\
 & DNN-RNT & 0.8383 & 0.8324 \\
 & GeneralAVQA & 0.9337 & 0.9380 \\
\hline

\end{tabular}
\caption{Performance (SROCC/PLCC) of baseline models on YT-NTU-AVQ dataset.}
\label{tab:avqa_baseline_results}
\end{table}

\begin{table}[t]
\centering
\renewcommand{\arraystretch}{1.1}
\setlength{\tabcolsep}{5pt}
\begin{tabular}{|c|c|c|c|c|}
\hline
\textbf{Group} & \textbf{$n$} & \textbf{Subj. A\%} & \textbf{Diff (V)} & \textbf{Diff (A)} \\
\hline
A $<<$ V  &  87  & 56.95 & -0.06 & +0.37 \\
A $<$  V  & 284  & 54.20 & +0.01 & +0.18 \\
A $\approx$ V  & 489  & 51.97 & +0.08 & +0.09 \\
A $>$  V  & 291  & 48.52 & +0.20 & +0.01 \\
A $>>$ V  & 469  & 45.08 & +0.34 & -0.17 \\
\hline
Overall   & 1620 & 50.01 & -0.02 & +0.05 \\
\hline
\end{tabular}
\caption{Subjective audio attention and AVQA–modality score differences across groups defined by the relative quality difference $AV\_AQA - AV\_VQA$ (thresholds: $\pm0.1$ slight, $\pm0.3$ large). \textbf{Subj. A\%}: mean subjective audio attention. \textbf{Diff (V/A)}: mean difference AVQA minus video/audio scores (positive indicates AVQA score is higher).}
\label{tab:modality_diff_groups}
\end{table}

\subsection{Analysis into Multimodal Quality Perception}

All three subjective scores (AVQA, AV\_VQA, AV\_AQA) show extremely high correlations, indicating that unimodal ratings are not strictly independent but largely shaped by overall perception. In particular, the near-perfect alignment between AVQA and AV\_VQA suggests that ``visual quality dominates'' in AVQA for in-the-wild UGC content. 

This is consistent with our baseline results (Table~\ref{tab:avqa_baseline_results}), VQA models achieve surprisingly high performance. We evaluate YT-NTU-AVQ using representative AQA, VQA, and AVQA models. Pretrained AQA models (PAM\cite{deshmukh2024pam}, AudioBox\cite{tjandra2025audiobox}) and VQA models (FasterVQA\cite{wu2023fastervqa}, FastVQA\cite{wu2022fastvqa}, Q-Align\cite{wu2023qlign}) are directly applied to the entire dataset. For AVQA, we evaluate fusion of A/V features by SVR \cite{min2020livesjtu} with representative feature combinations \cite{mittal2011brisque,davis1980mfcc,eyben2015eGeMAPS,saad2014vbliinds,cao2023sjtuuav} and deep learning-based models (DNN-RNT\cite{cao2023dnnrnt}, GeneralAVQA\cite{cao2023sjtuuav}) using 5-fold cross-validation.

A reasonable explanation is that audio quality in UGC is often concentrated in mid-to-high range with few severe distortions, and humans are less sensitive to subtle audio differences, making video variations appear more influential in quality assessment\cite{pinson2011avdominant,jin2022jnddmv,lin2022jndsurvey,mele2023roleaudio}. As a result, this apparent visual quality dominance is expected to be stronger in UGC datasets.

An interesting observation arises from the subjective attention weight. As shown in Table~\ref{tab:modality_diff_groups}, subjects report nearly balanced audio attention on average ($\mu = 50.01$ and $\sigma = 4.32$). However, when the quality scores of the two modalities differ, attention tends to shift toward the degraded modality, while the overall AVQA score remains closer to the higher-quality modality. This asymmetry suggests that humans focus on distortions but anchor their judgments by the better modality. Another important factor is semantic context: in scenarios where audio is important and highly correlated with video (music performance, dance, and people speaking), the reported audio attention is lower than in other scenarios, suggesting a counterintuitive holistic audio-visual integration.

\section{Conclusion}
We proposed the first crowdsourced subjective experiment framework tailored for AVQA, demonstrating that reliable quality assessment can be achieved even under non-laboratory settings. Combined with a designed data preparation strategy, this framework enabled the construction of YT-NTU-AVQ, the largest and most diverse AVQA dataset with multi-dimensional annotations. The dataset validates the feasibility of scaling AVQA beyond lab settings and provides a useful resource for future research as well as for understanding human multimodal quality perception.

\footnotesize
\bibliographystyle{IEEEbib}
\bibliography{refs}

@article{Akhtar2017avqasurvey,
  author  = {Z. Akhtar and T. H. Falk},
  title   = {Audio-Visual Multimedia Quality Assessment: A Comprehensive Survey},
  journal = {IEEE Access},
  volume  = {5},
  pages   = {21090--21117},
  year    = {2017},
  doi     = {10.1109/ACCESS.2017.2750918}
}

@article{zhang2025largemodelqa,
  author    = {Z. Zhang and Y. Zhou and C. Li and B. Zhao and X. Liu and G. Zhai},
  title     = {Quality Assessment in the Era of Large Models: A Survey},
  journal   = {ACM Transactions on Multimedia Computing, Communications, and Applications},
  volume    = {21},
  number    = {7},
  pages     = {189:1--189:31},
  year      = {2025},
  publisher = {ACM},
  doi       = {10.1145/3722559}
}

@article{wen2025cpllmcontextpixelaware,
  author  = {W. Wen and Y. Wu and Y. Sheng and N. Birkbeck and B. Adsumilli and Y. Wang},
  title   = {CP-LLM: Context and Pixel Aware Large Language Model for Video Quality Assessment},
  journal = {arXiv preprint arXiv:2505.16025},
  year    = {2025}
}

@techreport{itu_p910_2023,
  title        = {{ITU-T Recommendation P.910: Subjective Video Quality Assessment Methods for Multimedia Applications}},
  institution  = {International Telecommunication Union},
  number       = {T-REC-P.910-202310-I},
  year         = {2023},
  month        = {Oct.},
  note         = {Accessed: Jul. 28, 2025},
  url          = {https://www.itu.int/rec/T-REC-P.910-202310-I/en}
}

@article{martinez2014unbavq,
  author  = {H. B. Martinez and M. C. Q. Farias},
  title   = {Full-Reference Audio-Visual Video Quality Metric},
  journal = {J. Electron. Imaging},
  volume  = {23},
  number  = {6},
  pages   = {061108},
  year    = {2014}
}

@article{min2020livesjtu,
  author  = {X. Min and G. Zhai and J. Zhou and M. C. Q. Farias and A. C. Bovik},
  title   = {Study of Subjective and Objective Quality Assessment of Audio-Visual Signals},
  journal = {IEEE Trans. Image Process.},
  volume  = {29},
  pages   = {6054--6068},
  year    = {2020}
}

@inproceedings{cao2023sjtuuav,
  author    = {Y. Cao and X. Min and W. Sun and X. Zhang and G. Zhai},
  title     = {Audio-Visual Quality Assessment for User Generated Content: Database and Method},
  booktitle = {Proc. IEEE Int. Conf. Image Process. (ICIP)},
  pages     = {1495--1499},
  year      = {2023}
}

@inproceedings{demirbilek2016inrs,
  author    = {E. Demirbilek and J. C. Gr{\'e}goire},
  title     = {INRS Audiovisual Quality Dataset},
  booktitle = {Proc. ACM Int. Conf. Multimedia},
  pages     = {167--171},
  year      = {2016}
}

@misc{cdvl2019,
  author = {{Consumer Digital Video Library}},
  title  = {The Consumer Digital Video Library},
  year   = {2019},
  url    = {https://cdvl.org/}
}

@article{thomee2015yfcc100m,
  author  = {B. Thomee and D. A. Shamma and G. Friedland and B. Elizalde and K. Ni and D. Poland and D. Borth and L.-J. Li},
  title   = {The New Data and New Challenges in Multimedia Research},
  journal = {arXiv preprint arXiv:1503.01817},
  year    = {2015}
}

@inproceedings{hosu2017konstanz,
  author    = {V. Hosu and F. Hahn and M. Jenadeleh and H. Lin and H. Men and T. Szir{\'a}nyi and S. Li and D. Saupe},
  title     = {The Konstanz Natural Video Database (KoNViD-1k)},
  booktitle = {Proc. IEEE Int. Conf. Quality of Multimedia Experience (QoMEX)},
  pages     = {1--6},
  year      = {2017}
}

@article{sinno2018livevqc,
  author  = {Z. Sinno and A. C. Bovik},
  title   = {Large-Scale Study of Perceptual Video Quality},
  journal = {IEEE Trans. Image Process.},
  volume  = {28},
  number  = {2},
  pages   = {612--627},
  year    = {2018}
}

@inproceedings{wang2019youtubeugc,
  author    = {Y. Wang and S. Inguva and B. Adsumilli},
  title     = {YouTube UGC Dataset for Video Compression Research},
  booktitle = {Proc. IEEE Int. Workshop Multimedia Signal Process. (MMSP)},
  pages     = {1--5},
  year      = {2019}
}

@inproceedings{ying2021lsvq,
  author    = {Z. Ying and M. Mandal and D. Ghadiyaram and A. C. Bovik},
  title     = {Patch-VQ: {``Patching Up''} the Video Quality Problem},
  booktitle = {Proc. IEEE/CVF Conf. Comput. Vis. Pattern Recognit. (CVPR)},
  pages     = {14019--14029},
  year      = {2021}
}

@article{lin2022konjnd,
  author  = {H. Lin and G. Chen and M. Jenadeleh and V. Hosu and U.-D. Reips and R. Hamzaoui and D. Saupe},
  title   = {Large-Scale Crowdsourced Subjective Assessment of Picturewise Just Noticeable Difference},
  journal = {IEEE Trans. Circuits Syst. Video Technol.},
  volume  = {32},
  number  = {9},
  pages   = {5859--5873},
  year    = {2022}
}

@article{chen2023valor,
  author  = {S. Chen and X. He and L. Guo and X. Zhu and W. Wang and J. Tang and J. Liu},
  title   = {VALOR: Vision-Audio-Language Omni-Perception Pretraining Model and Dataset},
  journal = {arXiv preprint arXiv:2304.08345},
  year    = {2023}
}

@inproceedings{wang2024youtube,
  author    = {Y. Wang and J. G. Yim and N. Birkbeck and B. Adsumilli},
  title     = {YouTube SFV+ HDR Quality Dataset},
  booktitle = {Proc. IEEE Int. Conf. Image Process. (ICIP)},
  pages     = {96--102},
  year      = {2024}
}

@article{chen2025mugsqa,
  author  = {T. Chen and J. Jin and S. Cai and Z. Li and W. Lin},
  title   = {MUGSQA: Novel Multi-Uncertainty-Based Gaussian Splatting Quality Assessment Method, Dataset, and Benchmarks},
  journal = {arXiv preprint arXiv:2511.06830},
  year    = {2025}
}

@article{tjandra2025audiobox,
  author  = {A. Tjandra and Y.-C. Wu and B. Guo and J. Hoffman and B. Ellis and A. Vyas and B. Shi and S. Chen and M. Le and N. Zacharov and others},
  title   = {Meta AudioBox Aesthetics: Unified Automatic Quality Assessment for Speech, Music, and Sound},
  journal = {arXiv preprint arXiv:2502.05139},
  year    = {2025}
}

@article{wu2023fastervqa,
  author  = {H. Wu and C. Chen and L. Liao and J. Hou and W. Sun and Q. Yan and J. Gu and W. Lin},
  title   = {Neighbourhood Representative Sampling for Efficient End-to-End Video Quality Assessment},
  journal = {IEEE Trans. Pattern Anal. Mach. Intell.},
  volume  = {45},
  number  = {12},
  pages   = {15185--15202},
  year    = {2023}
}

@inproceedings{gemmeke2017audioset,
  title={Audio set: An ontology and human-labeled dataset for audio events},
  author={J. F. Gemmeke and D. P. W. Ellis and D. Freedman and A. Jansen and W. Lawrence and R. C. Moore and M. Plakal and M. Ritter},
  booktitle={2017 IEEE international conference on acoustics, speech and signal processing (ICASSP)},
  pages={776--780},
  year={2017},
  organization={IEEE}
}

@article{de2015jspsych,
  author  = {J. R. De Leeuw},
  title   = {jsPsych: A JavaScript Library for Creating Behavioral Experiments in a Web Browser},
  journal = {Behav. Res. Methods},
  volume  = {47},
  number  = {1},
  pages   = {1--12},
  year    = {2015}
}

@article{deshmukh2024pam,
  title={Pam: Prompting audio-language models for audio quality assessment},
  author={S. Deshmukh and D. Alharthi and B. Elizalde and H. Gamper and M. A. Ismail and R. Singh and B. Raj and H. Wang},
  journal={arXiv preprint arXiv:2402.00282},
  year={2024}
}

@inproceedings{wu2022fastvqa,
  author    = {H. Wu and C. Chen and J. Hou and L. Liao and A. Wang and W. Sun and Q. Yan and W. Lin},
  title     = {Fast-VQA: Efficient End-to-End Video Quality Assessment with Fragment Sampling},
  booktitle = {Proc. Eur. Conf. Comput. Vis. (ECCV)},
  pages     = {538--554},
  year      = {2022}
}

@article{wu2023qlign,
  author  = {H. Wu and Z. Zhang and W. Zhang and C. Chen and L. Liao and C. Li and Y. Gao and A. Wang and E. Zhang and W. Sun and others},
  title   = {Q-Align: Teaching LMMs for Visual Scoring via Discrete Text-Defined Levels},
  journal = {arXiv preprint arXiv:2312.17090},
  year    = {2023}
}

@inproceedings{mittal2011brisque,
  author    = {A. Mittal and A. K. Moorthy and A. C. Bovik},
  title     = {Blind/Referenceless Image Spatial Quality Evaluator},
  booktitle = {Proc. Asilomar Conf. Signals, Systems, Computers},
  pages     = {723--727},
  year      = {2011}
}

@ARTICLE{davis1980mfcc,
  author  = {Davis, S. and Mermelstein, P.},
  journal = {IEEE Transactions on Acoustics, Speech, and Signal Processing}, 
  title   = {Comparison of parametric representations for monosyllabic word recognition in continuously spoken sentences}, 
  year    = {1980},
  volume  = {28},
  number  = {4},
  pages   = {357-366},
  doi     = {10.1109/TASSP.1980.1163420}}

@article{eyben2015eGeMAPS,
  author  = {F. Eyben and K. R. Scherer and B. W. Schuller and J. Sundberg and E. Andr{\'e} and C. Busso and L. Y. Devillers and J. Epps and P. Laukka and S. S. Narayanan and others},
  title   = {The Geneva Minimalistic Acoustic Parameter Set (GeMAPS) for Voice Research and Affective Computing},
  journal = {IEEE Trans. Affect. Comput.},
  volume  = {7},
  number  = {2},
  pages   = {190--202},
  year    = {2015}
}

@article{saad2014vbliinds,
  author  = {M. A. Saad and A. C. Bovik and C. Charrier},
  title   = {Blind Prediction of Natural Video Quality},
  journal = {IEEE Trans. Image Process.},
  volume  = {23},
  number  = {3},
  pages   = {1352--1365},
  year    = {2014}
}

@ARTICLE{cao2023dnnrnt,
  author  = {Y. Cao and X. Min and W. Sun and G. Zhai},
  journal = {IEEE Transactions on Image Processing}, 
  title   = {Attention-Guided Neural Networks for Full-Reference and No-Reference Audio-Visual Quality Assessment}, 
  year    = {2023},
  volume  = {32},
  number  = {},
  pages   = {1882-1896},
  doi     = {10.1109/TIP.2023.3251695}}

@article{lin2022jndsurvey,
  author  = {W. Lin and G. Ghinea},
  title   = {Progress and Opportunities in Modelling Just-Noticeable Difference (JND) for Multimedia},
  journal = {IEEE Transactions on Multimedia},
  year    = {2022},
  volume  = {24},
  pages   = {3706--3721},
  doi     = {10.1109/TMM.2021.3106503}
}

@article{jin2022jnddmv,
  author  = {J. Jin and X. Zhang and X. Fu and H. Zhang and W. Lin and J. Lou and Y. Zhao},
  title   = {Just Noticeable Difference for Deep Machine Vision},
  journal = {IEEE Transactions on Circuits and Systems for Video Technology},
  year    = {2022},
  volume  = {32},
  number  = {6},
  pages   = {3452--3461},
  doi     = {10.1109/TCSVT.2021.3113572}
}

@inproceedings{mele2023roleaudio,
  author    = {M. L. Mele and D. Millar and S. Colabrese},
  title     = {The Role of Audio in Visual Perception of Quality},
  booktitle = {HCI International 2023 – Late Breaking Papers: 25th International Conference on Human-Computer Interaction (HCII)},
  pages     = {155-167},
  year      = {2023},
  doi       = {10.1007/978-3-031-48038-6_10}
}

@ARTICLE{pinson2011avdominant,
  author={M. H. Pinson and L. Janowski and R. P{\'e}pion and Q. Huynh-Thu and C. Schmidmer and P. Corriveau and A. Younkin and P. Le Callet and M. Barkowsky and W. Ingram},
  journal={IEEE Journal of Selected Topics in Signal Processing}, 
  title={The Influence of Subjects and Environment on Audiovisual Subjective Tests: An International Study}, 
  year={2012},
  volume={6},
  number={6},
  pages={640-651},
  doi={10.1109/JSTSP.2012.2215306}}

@article{comanici2025gemini2.5,
  author  = {G. Comanici and E. Bieber and M. Schaekermann and I. Pasupat and N. Sachdeva and I. Dhillon and M. Blistein and O. Ram and D. Zhang and E. Rosen and others},
  title   = {Gemini 2.5: Pushing the Frontier with Advanced Reasoning, Multimodality, Long Context, and Next Generation Agentic Capabilities},
  journal = {arXiv preprint arXiv:2507.06261},
  year    = {2025}
}

\clearpage
\onecolumn
\appendix

\begin{center}
    {\LARGE \bf Supplementary Material}\\[1em]
\end{center}

\vspace{1.5em}

\begingroup
\normalsize

This supplementary material focuses on implementation details of the experimental platform, crowdsourcing setup, and dataset construction.
It is intended to support reproducibility and facilitate future dataset extension or similar experiment design.

\section{Experimental Platform}
\subsection{Environment Check}

Before starting the rating session, subjects are guided through an environment preparation and hardware verification stage. In web-based crowdsourcing experiments, strict control over the testing environment is not feasible. Therefore, we provide clear preparation instructions and perform lightweight automatic checks to reduce the risk of unreliable experimental conditions.

Subjects are first asked to confirm recommended conditions (Figure~\ref{fig:env_confirm}), including a quiet environment, use of a desktop device with a sufficiently large screen, headphone-based listening, and stable network connectivity. If the subject indicates that the environment is not ready, the experiment cannot proceed.

\begin{figure}[h]
  \centering
  \includegraphics[width=0.7\linewidth]{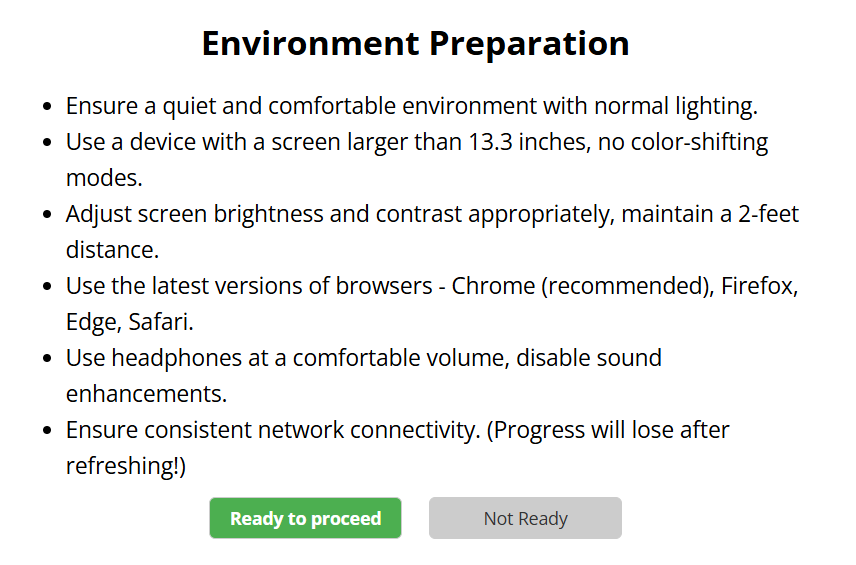}
  \caption{Screenshot of the Environment Preparation Confirmation Page}
  \label{fig:env_confirm}
\end{figure}

The platform then performs basic automatic checks using browser APIs, including:

\begin{itemize}
    \item \textbf{Display capability:} verifying that the effective screen resolution is larger than 720p. (Enforcing a strict 1080p requirement is unreliable in web environments due to variations in browser reporting, device pixel ratio, and operating system scaling)
    \item \textbf{Network connectivity:} performing a short media download test to ensure stable video loading.
    \item \textbf{Audio output availability:} verifying that an audio output device is present.
\end{itemize}

If any check fails, subjects are allowed a limited number of retries. If all retries fail, the experiment session is terminated. 

In addition, the User-Agent string is recorded during task submission to help identify the client device type and to detect submissions from unsupported device categories (e.g., mobile devices).

\subsection{A/V Sequence Rating Interface}

\begin{figure}[h]
  \centering
  \includegraphics[width=0.7\linewidth]{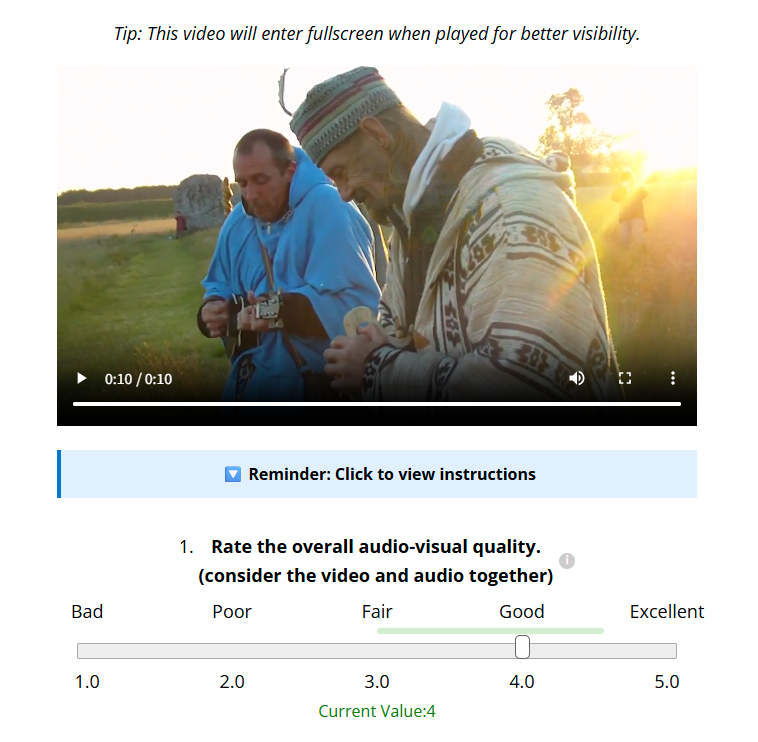}
  \caption{Screenshot of the A/V Sequence Rating Interface}
  \label{fig:avqa_interface_cut1}
\end{figure}

In each experimental session, subjects consecutively rate 30 A/V sequences. Upon entering the page shown in Figure \ref{fig:avqa_interface_cut1}, the video automatically starts in full-screen mode. During the first playback, any attempt to drag the progress bar triggers a warning, and submission is not allowed unless the total watch time meets or exceeds the video duration. After the initial playback, subjects may freely replay the video as many times as they wish. Throughout the experiment, each video is preloaded to minimize network-induced stuttering.

Following each video, subjects provide ratings using a single-stimulus evaluation interface with four slider-based questions. The primary rating question asks for an overall audio-visual quality score on a continuous scale from 1 to 5 with a step size of 0.1. The slider initially appears at the leftmost position and must be moved before submission is enabled.

The interface also provides a collapsible Instruction panel and on-hover descriptions for each question. During the training phase, an expected score range (displayed as a green bar) is shown as a broad guideline; subjects are required to adjust their sliders within this range to proceed, ensuring they become familiar with the approximate rating standard.

\begin{figure}[h]
  \centering
  \includegraphics[width=0.7\linewidth]{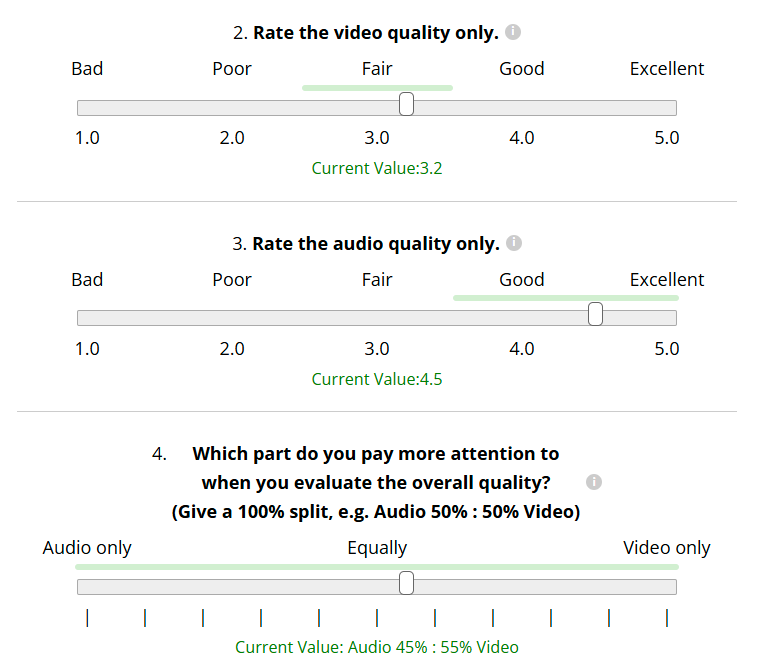}
  \caption{Screenshot of the Additional Questions}
  \label{fig:avqa_interface_cut2}
\end{figure}

To further explore the mechanisms underlying human perception of audio-visual quality, three supplementary questions (Q2–Q4) were added to the interface, all answered via slider controls. Q2 and Q3 use the same 1–5 scale as Q1, while Q4 captures subjects’ perceived attention weights between audio and video, represented as a continuous allocation ranging from 0\%:100\% (audio:video) to 100\%:0\%, with a step size of 1\%.

\subsection{Task Assignment and Submission}

We deployed a custom backend server to manage task distribution and data collection. The subject’s AMT worker ID is transmitted to the server when requesting a task, allowing the system to check submission history and prevent duplicate task assignments. All experimental videos are pre-grouped in advance. When a subject requests a task, the server randomly assigns one unfinished task group based on the subject’s task history.

After completing the experiment, all rating results are transmitted to the backend server and stored for subsequent statistical analysis. Upon successful submission, the server generates a random completion code, which is returned to the subject for confirmation of task completion on the AMT platform. 

This design ensures balanced task distribution across video groups and reduces the risk of repeated submissions, preventing certain videos from being oversampled and helping maintain a more uniform number of ratings per video.

\section{Crowdsourcing Experiment}
\subsection{Qualification Settings}
To ensure high-quality participation, we adopted a multi-stage qualification strategy. The Formal Test stage was restricted to experienced and reliable workers who satisfied all predefined criteria.

\begin{figure}[h]
\centering
\includegraphics[width=0.85\linewidth]{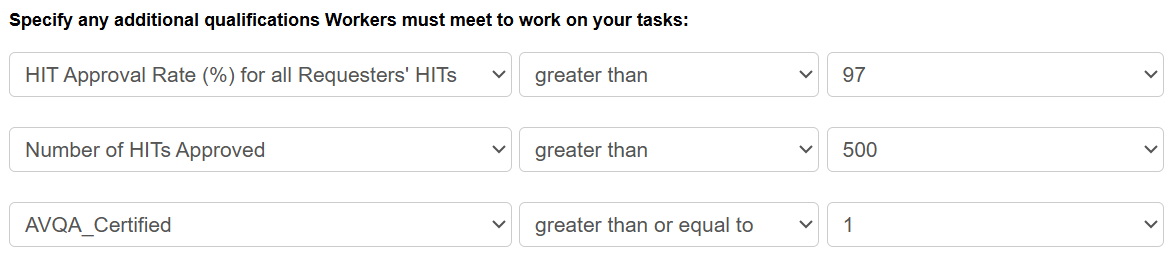}
\caption{Qualification requirements for workers to participate in the Formal Test on AMT.}
\label{fig:qualification_setting}
\end{figure}

As shown in Figure~\ref{fig:qualification_setting}, three qualification filters were enforced for the Formal Test:

\textbf{HIT Approval Rate:} Workers were required to have an approval rate above 97\% for all previously completed tasks.

\textbf{Number of HITs Approved:} Workers needed to have successfully completed more than 500 tasks.

\textbf{AVQA\_Certified Qualification:} This custom qualification was only granted to workers whose submissions in both the Pretest and Qualification Test stages passed the dynamic filtering procedure described in Section~B.2.

Here, a HIT (Human Intelligence Task) refers to an individual task posted on the AMT platform. Each HIT in our study corresponds to a single experimental session where the subject rates a set of videos.

To streamline qualification management, we leveraged the AMT API to automatically update qualifications in bulk. Upon verifying a worker’s eligibility, our backend system issued the AVQA\_Certified flag programmatically, ensuring that only qualified subjects could access the Formal Test without manual intervention.

\subsection{Dynamic Filtering of Subjective Data}
To ensure the reliability of crowdsourced ratings, we designed a dynamic filtering method based on rank correlation and score variance. First, all invalid submissions were removed. We then aggregated the ratings of all valid submissions to compute the Mean Opinion Score (MOS) for each video. Using these preliminary MOS values as the reference, we evaluated the quality of each individual submission by calculating:

\textbf{SROCC (Spearman Rank-Order Correlation Coefficient)} between the subject’s ratings and the aggregated MOS, computed separately for the three rating dimensions (AVQA, AV\_VQA, AV\_AQA) and then averaged.

\textbf{STD (Standard Deviation)} of the subject’s ratings across the three rating dimensions, also averaged.

Submissions that achieved both an average SROCC and average STD above the thresholds (0.5, 0.5) were retained, and the MOS was recomputed using only the filtered ratings to obtain the final subjective scores.

\begin{verbatim}
Algorithm: Dynamic Filtering of Crowdsourced Ratings
1. Remove duplicate and invalid submissions.
2. Compute preliminary MOS for each video.
3. For each submission s:
   a. Merge s with video MOS labels.
   b. Calculate SROCC(s, MOS) for AVQA, AV_VQA, AV_AQA.
   c. Calculate STD(s) for AVQA, AV_VQA, AV_AQA.
   d. Compute avg_SROCC and avg_STD.
4. Keep s if avg_SROCC > 0.5 and avg_STD > 0.5.
5. Recompute final MOS using only valid ratings.
\end{verbatim}

In principle, preliminary MOS can be recomputed either after each round of filtering or when new crowdsourced data are added. However, in practice we observe that MOS estimates remain relatively stable, and therefore iterative refinement is not necessary.

The above filtering procedure was applied consistently across all stages of the crowdsourcing experiment. The resulting data retention statistics for each stage are summarized below.

\textbf{Pretest.} We released 328 tasks for 120 videos (4 groups). From 286 subjects, 9,840 ratings were collected. After dynamic filtering, 125 valid tasks remained, producing 3,750 reliable ratings (31.25 per video, 38.1\% acceptance). These scores served as initial references for qualification.

\textbf{Qualification Test.} 1,952 new subjects rated one pretest group each. Dynamic filtering retained 958 (49.08\%), and together with 93 pretest-qualified subjects, 1,051 subjects were qualified. Final MOS for the 120 videos was recomputed using 32,490 accepted ratings (270.75 per video).

\textbf{Formal Test.} For the remaining 1,500 videos, 4,680 tasks were launched. From 580 qualified subjects, 3,153 tasks passed filtering, yielding 94,590 reliable ratings (63.06 per video, 67.37\% acceptance).

\section{Dataset Details}
\subsection{Sampling Strategy}

To construct a representative subset under limited dataset size, we sample sequences to balance key audio-visual attributes while preserving real-world UGC characteristics. Although many attributes (e.g., perceptual quality, low-level signal statistics, semantic categories) can be considered, many are intrinsically correlated. Enforcing uniform coverage across all attributes is impractical due to data sparsity. Therefore, we prioritize three key attributes with a focus on audio to support deeper analysis.

\textbf{Audio quality.} We use AudioBox to generate pseudo-labels and adopt its Clarity Estimation (CE) and Perceived Quality (PQ) metrics to capture perceptual degradations across diverse audio types.

\textbf{Video quality.} We adopt FasterVQA to provide reliable video quality predictions while maintaining computational efficiency.

\textbf{Audio semantic category.} We map AudioSet labels into three coarse categories (speech, music, sound) and consider their seven label combinations to improve semantic coverage.

We adopted a stratified sampling approach that jointly considers audio category combinations and pseudo-quality scores for audio and video. Discrete audio labels were balanced according to predefined ratios (with single categories weighted as 2 and combined categories weighted as 1), while continuous features (pseudo-quality) were softly equalized by binning and applying a smoothing coefficient. This process slightly increases the sampling probability of underrepresented bins without heavily distorting the original distribution. The balancing strength was controlled by a coefficient $\alpha = 0.3$, 
and continuous features were discretized into $n = 8$ bins. 
The parameter $\alpha$ adjusts the degree of equalization: when $\alpha = 0$, no balancing is applied; larger $\alpha$ values lead to stronger flattening of the bin distribution.

\begin{verbatim}
Algorithm: Stratified Sampling with Soft Balancing
1. Define target ratios for discrete audio label groups.
2. For each group:
   a. Discretize continuous pseudo-quality features into n_bins (n=8).
   b. For each bin i, compute original probability p_i.
   c. Define uniform distribution u_i = 1 / n_bins.
   d. Calculate bin weight w_i = (u_i / p_i)^alpha, normalize to sum=1.
   e. Assign each sample a weight according to its bin's w_i.
   f. Sample target_n videos proportionally to the adjusted weights.
3. Merge all sampled groups to form a balanced set of 10,000 videos.
4. Randomly select 1,296 videos from the 10,000 for the final dataset.
\end{verbatim}

\begin{figure}[t]
    \centering
    \includegraphics[width=0.8\linewidth]{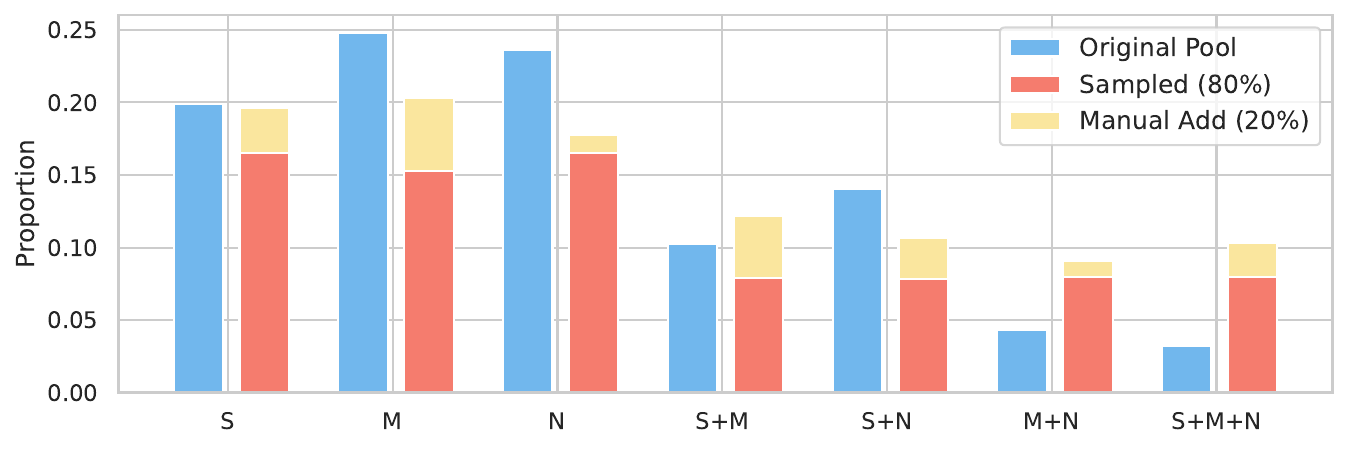} 
    \caption{Distribution of audio semantic label combinations after sampling.}
    \label{fig:Audio_Category_Distribution}
\end{figure}

\begin{figure}[t]
  \centering
  \includegraphics[width=0.6\linewidth]{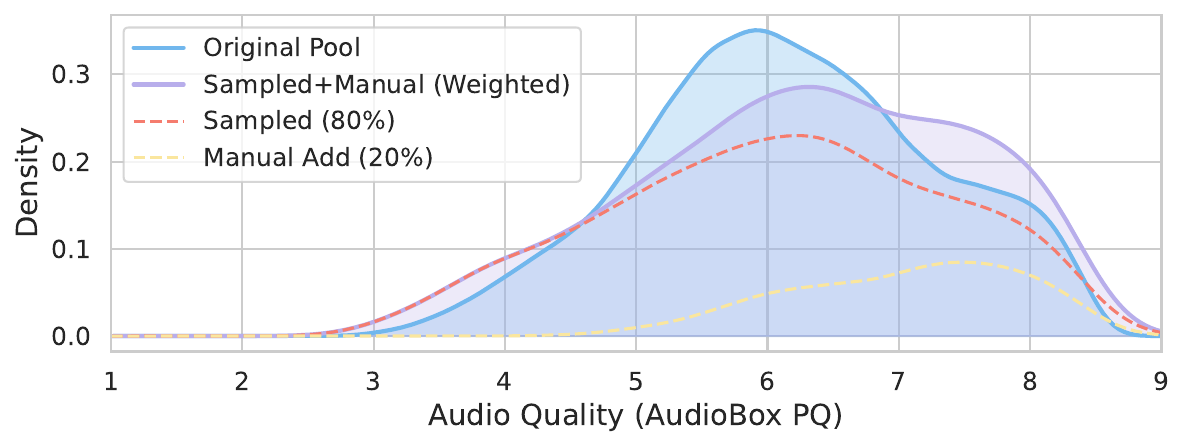}
  \vspace{4pt}

  \includegraphics[width=0.6\linewidth]{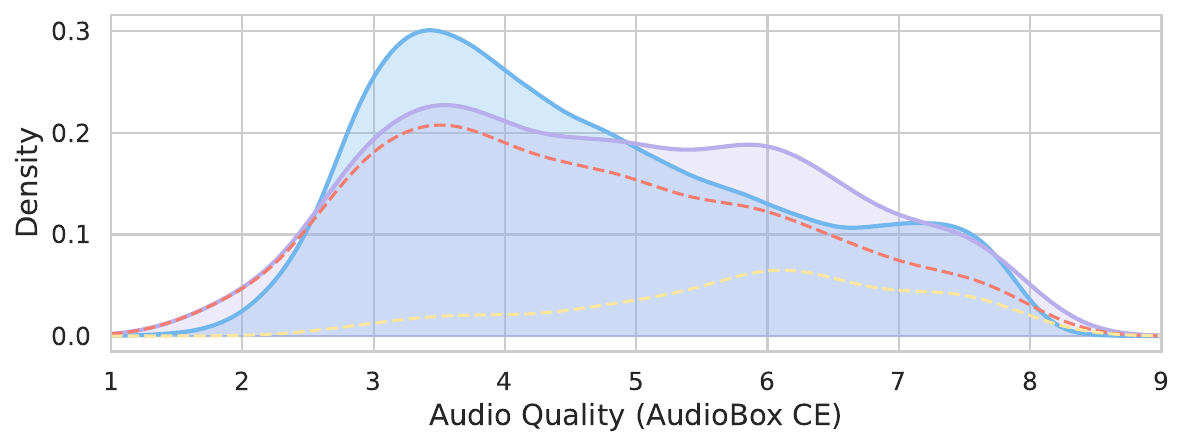}
  \vspace{4pt}

  \includegraphics[width=0.6\linewidth]{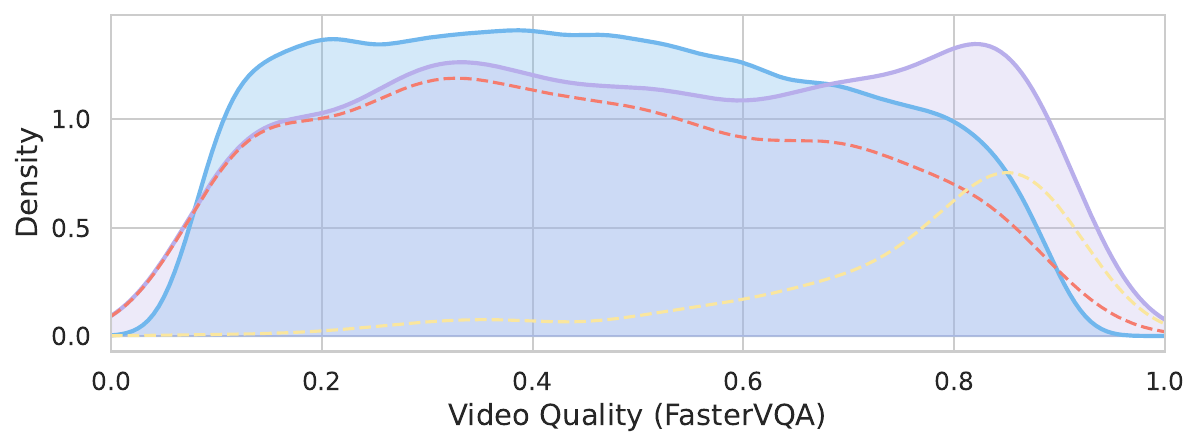}

  \caption{Distributions of pseudo labels before and after sampling. From top to bottom: audio quality (AudioBox PQ), audio quality (AudioBox CE), and video quality (FasterVQA). Weighted sampling and manual additions improve label coverage across the score range.}
  \label{fig:label_distributions}
\end{figure}

Using this approach, we first sampled 10,000 videos from the VALOR-1M pool with improved coverage of labels and quality ranges. From this set, 1,296 videos were randomly selected as the final dataset for subjective experiments.

The resulting distribution of audio semantic combinations after sampling is shown in Figure~\ref{fig:Audio_Category_Distribution}.

The distributions of pseudo labels before and after sampling are shown in Figure~\ref{fig:label_distributions}, including audio quality based on AudioBox predictions, and video quality based on FasterVQA predictions.

\subsection{Manual Search Procedure}
To complement automated sampling, we performed manual video selection to ensure diversity and content quality. Following the category design of prior datasets such as YouTube SFV+HDR, we defined ten categories: \textbf{Animal, Cooking, Dance, Gameplay, Health, Hobby, Music, Travel, Speech}, and \textbf{Sports}. For each category, more than thirty keywords were generated with the assistance of ChatGPT and manually refined to guide the search process.

Candidate videos were retrieved from YouTube under Creative Commons licenses and manually screened to ensure:
\begin{itemize}
    \item relevance to the intended category with diverse scenes and creators;
    \item sufficient visual quality and recent upload time (preferably within the last two years);
    \item landscape orientation and appropriate content;
    \item exclusion of unsuitable material, such as re-uploaded or disturbing content.
\end{itemize}

For each keyword, at least the top 50 search results were reviewed, and eligible videos were selected to supplement the final pool.

\subsection{Dataset Diversity Analysis}

To further evaluate dataset diversity, we analyze multiple complementary aspects including low-level signal attributes, duration distribution, resolution distribution, and semantic category coverage.

\begin{figure}[t]
  \centering

  \subfloat[Temporal Information (TI)]{
    \includegraphics[width=0.48\linewidth]{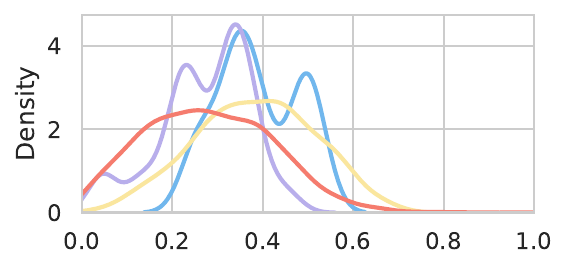}
  }
  \hfill
  \subfloat[Spatial Information (SI)]{
    \includegraphics[width=0.48\linewidth]{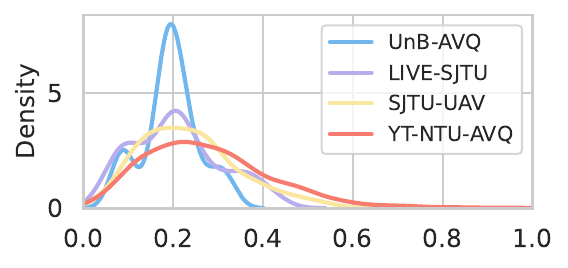}
  }

  \medskip

  \subfloat[Spectral Centroid (SC)]{
    \includegraphics[width=0.48\linewidth]{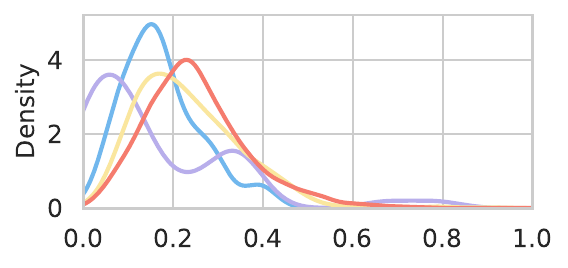}
  }
  \hfill
  \subfloat[Spectral Entropy (SE)]{
    \includegraphics[width=0.48\linewidth]{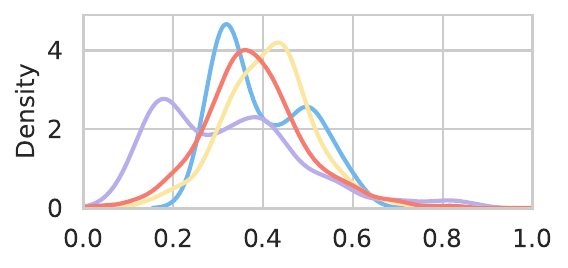}
  }

  \caption{Distributions of normalized low-level audio-visual attributes across four datasets: UnB-AVQ, LIVE-SJTU, SJTU-UAV, and our YT-NTU-AVQ. Our dataset demonstrates broader coverage in both video (TI, SI) and audio (SC, SE) domains.}
  \label{fig:avqa_attributes}
\end{figure}

\textbf{(1) Low-level audio-visual attributes.}

To further evaluate dataset diversity, we analyzed several low-level audio-visual attributes and compared their distributions with representative AVQA datasets. Figure~\ref{fig:avqa_attributes} illustrates four representative attributes, including video spatial information (SI), temporal information (TI), and audio spectral centroid (SC) and spectral entropy (SE).

Although these attributes were not explicitly controlled during sampling, our dataset maintains distributions comparable to the UGC-based dataset SJTU-UAV, indicating good coverage of low-level audio-visual characteristics. Compared with earlier AVQA datasets, our dataset also exhibits broader coverage across both video (TI, SI) and audio (SC, SE) dimensions.

\textbf{(2) Duration consistency.}

To facilitate processing and evaluation, A/V sequences are extracted with a target duration of 10 seconds, consistent with AudioSet. Due to source video boundaries and trimming constraints, a small portion of clips deviate slightly from the target duration. In practice, the average duration is 9.91 seconds with a standard deviation of 0.59 seconds, and 92.5\% of clips fall within 9.8–10.2 seconds. This high consistency ensures stable subjective evaluation and model training.

\textbf{(3) Resolution distribution.}

The dataset contains a wide range of spatial resolutions, including many non-standard formats commonly observed in real-world UGC content, reflecting realistic diversity in capture devices and recording settings. The spatial resolution distribution is illustrated in Figure~\ref{fig:resolution_distribution}.

\begin{figure}[t]
  \centering
  \includegraphics[width=0.9\linewidth]{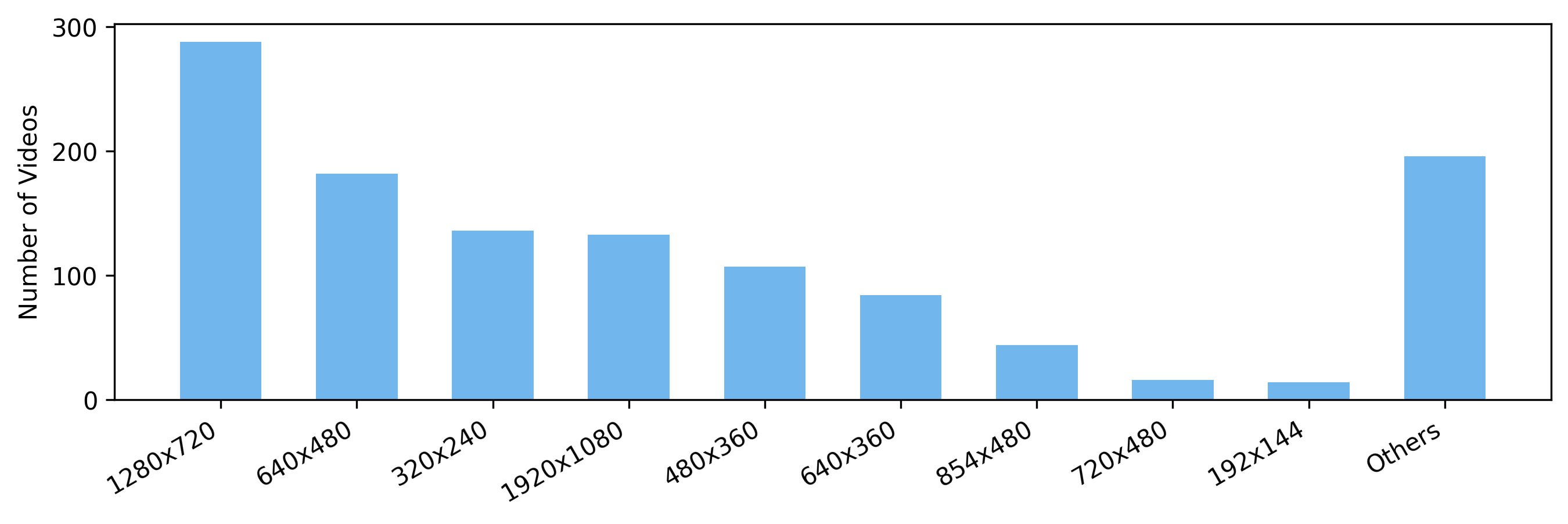}
  \caption{Distribution of spatial resolutions in YT-NTU-AVQ. The most frequent resolutions are shown individually, while the remaining resolutions are aggregated into an ``Others'' category.}
  \label{fig:resolution_distribution}
\end{figure}

\begin{figure}[t]
  \centering
  \includegraphics[width=0.9\linewidth]{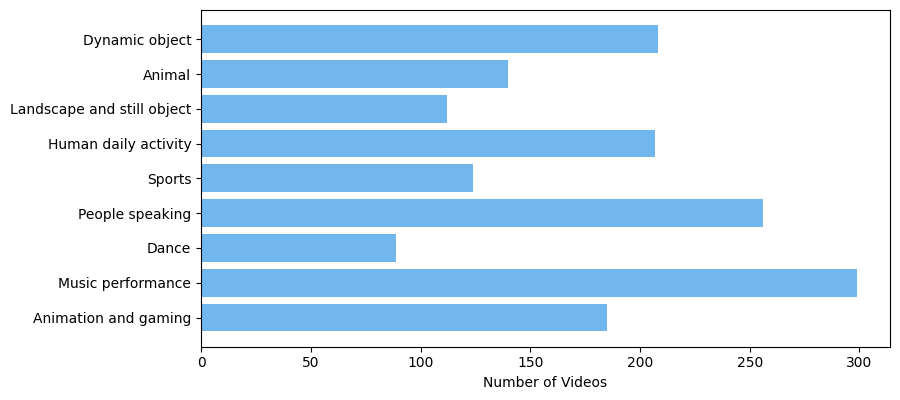}
  \caption{Distribution of videos across predefined coarse semantic content categories in YT-NTU-AVQ.}
  \label{fig:semantic_distribution}
\end{figure}

\textbf{(4) Semantic category distribution.}

To analyze content scenario diversity, we define 9 coarse semantic video content categories based on dominant visual and contextual content. These categories are designed to reflect common audio-visual interaction scenarios rather than fine-grained semantic taxonomy. Video-level labels are assigned using Gemini 2.5 based on primary content description.

The semantic category distribution is shown in Figure~\ref{fig:semantic_distribution}.

\endgroup

\end{document}